\documentclass[12pt,letterpaper]{article}
\usepackage[a4paper, total={7in, 10in}]{geometry}

\usepackage{graphicx}
\usepackage{helvet}
\usepackage{authblk}
\usepackage{hyperref}
\usepackage{amsmath} 
\usepackage{amssymb} 
\usepackage{orcidlink} 
\usepackage[super,comma,sort&compress]  
   {natbib}
\usepackage[right]{lineno} \linenumbers
\usepackage[title]{appendix}%

\usepackage[T1]{fontenc}    
\usepackage{url}            
\usepackage{booktabs}       
\usepackage{amsfonts}       
\usepackage{nicefrac}       
\usepackage{microtype}      
\usepackage{xcolor}         
\usepackage{multirow}
\usepackage{threeparttable}
\usepackage{amsmath}
\usepackage{subcaption}
\usepackage{makecell}
\usepackage{placeins}
\usepackage{booktabs}      
\usepackage{siunitx}       
\usepackage{caption}       
\usepackage{tabularx}
\usepackage{listings}    
\usepackage{pifont}
\usepackage{array}
\usepackage{makecell}
\usepackage{svg}
\theadset{\bfseries}

\makeatletter
\renewcommand{\maketitle}{\bgroup\setlength{\parindent}{0pt}
\begin{flushleft}
  \textbf{\@title}
  
  \@author
\end{flushleft}\egroup}
\makeatother

\title{Artificial Intelligence-Enabled Spirometry for Early Detection of Right Heart Failure}

\date{}

\author[1,5,\#]{Bin Liu \orcidlink{0009-0001-1588-6487}}
\author[3,\#]{Qinghao Zhao}
\author[1,2,*]{Yuxi Zhou}
\author[1,5]{Zhejun Sun}
\author[1]{Kaijie Lei}
\author[4]{Deyun Zhang}
\author[4]{Shijia Geng}
\author[5,6,7,8,*]{Shenda Hong \orcidlink{0000-0001-7521-5127}}

\affil[1]{Department of Computer Science, Tianjin University of Technology, Tianjin, China}
\affil[2]{DCST, BNRist, RIIT, Institute of Internet Industry, Tsinghua University, Beijing, China}
\affil[3]{Department of Cardiology, Peking University People's Hospital, Beijing, China}
\affil[4]{HeartVoice Medical Technology, Hefei, China}
\affil[5]{National Institute of Health Data Science, Peking University, Beijing, China}
\affil[6]{State Key Laboratory of Vascular Homeostasis and Remodeling, NHC Key Laboratory of Cardiovascular Molecular Biology and Regulatory Peptides, Peking University, Beijing, China}
\affil[7]{Institute of Medical Technology, Peking University Health Science Center, Beijing, China}
\affil[8]{Institute for Artificial Intelligence, Peking University, Beijing, China}

\affil[$\#$]{These authors contributed equally}
\affil[*]{Correspondence: joy\_yuxi@pku.edu.cn, hongshenda@pku.edu.cn}

\begin{document}

\maketitle

\section*{ABSTRACT}
Right heart failure (RHF) is a disease characterized by abnormalities in the structure or function of the right ventricle (RV), which is associated with high morbidity and mortality. Lung disease often causes increased right ventricular load, leading to RHF. Therefore, it is very important to screen out patients with cor pulmonale who develop RHF from people with underlying lung diseases. In this work, we propose a self-supervised representation learning method to early detecting RHF from patients with cor pulmonale, which uses spirogram time series to predict patients with RHF at an early stage. The proposed model is divided into two stages. The first stage is the self-supervised representation learning-based spirogram embedding (SLSE) network training process, where the encoder of the Variational autoencoder (VAE-encoder) learns a robust low-dimensional representation of the spirogram time series from the data-augmented unlabeled data. The second stage is the downstream classification task, where the low-dimensional representation of the spirogram extracted by the VAE-encoder is fused with demographic information as the input to the CatBoost classifier, and finally predicts patients with RHF. After training and testing on the UK Biobank, which includes a carefully selected subset of 26,617 individuals suitable for the study, our model achieved an AUROC of 0.7501 in detecting patients with RHF, demonstrating a strong ability to distinguish different subsets within the population. In addition, chronic kidney disease (CKD) and valvular heart disease (VHD) are known risk factors for RHF, we further evaluate it on a test set consisting exclusively of 74 patients with CKD and another test set consisting exclusively of 64 patients with VHD. The AUROC values of the model on these two subsets were 0.8194 and 0.8413, respectively. This result highlights the model’s potential utility in prediction RHF among patients with CKD and VHD, populations with elevated clinical risk. In conclusion, this study presents a self-supervised representation learning approach that leverages spirogram time series data combined with demographic information to detect RHF and demonstrate promising potential for early RHF detection in clinical practice.

\section*{KEYWORDS}
Right heart failure (RHF), self-supervised representation learning, spirogram

\section*{INTRODUCTION}

Right heart failure (RHF) is caused by structural or functional abnormalities of the right ventricle (RV) due to pressure overload, volume overload, or intrinsic myocardial contractile dysfunction, and is associated with a range of clinical syndromes with high morbidity and mortality \cite{maraey}. This condition is observed in \verb|3% to 9%| of patients admitted for acute heart failure, and the in-hospital mortality rate for patients with acute RHF is \verb|5% to 17%| \cite{nieminen}. Therefore, early diagnosis and intervention are crucial for the prognosis of patients with RHF. \cite{marvin}.

Chronic lung diseases, including chronic obstructive pulmonary disease (COPD), interstitial lung disease, and pulmonary vascular disease, commonly increase the afterload on the right ventricle (RV). This intensifies the burden on the right heart, ultimately leading to right heart failure (RHF) \cite{rosenkranz}. For patients with lung disease complicated by RHF, early detection and prompt initiation of anti-heart failure interventions are crucial for improving outcomes \cite{magnussen}. Currently, echocardiography, cardiac magnetic resonance imaging, and pulmonary artery catheterization are the primary methods for assessing RV function \cite{marvin}. However, the progression of RV dysfunction in patients with lung disease is often subtle, and routine cardiac function assessments, such as echocardiography, are not typically conducted in this patient population. This frequently leads to delayed or even missed diagnoses of RHF, thereby accelerating the deterioration of patient prognosis.

As a routine examination for lung diseases, spirometry is regularly performed in patients with chronic lung disease to assess pulmonary function. Studies have shown that patients with heart failure may exhibit specific characteristic changes in pulmonary function, such as reduced Forced Vital Capacity (FVC) and an obstructive-like pattern (decreased FEV1/FVC) \cite{magnussen}. However, these characteristics are too subtle and complex, making them difficult to distinguish from the spirometric characteristics of lung diseases. 

In recent years, with the advancement of deep learning technology, its powerful capabilities in feature extraction and pattern recognition have shown immense potential in disease diagnosis \cite{esteva, yuanyuan}. By thoroughly analyzing spirogram data, deep learning can accurately identify complex feature associations related to RHF that may be overlooked through manual observation, thus offering a feasible approach to detecting RHF in patients based on spirograms.

In addition, the acquisition of spirograms depends on clinical spirometry tests and existing curve generation technology \cite{cosentino}. The former often results in poor results of spirometry tests due to the use of equipment with different sampling frequencies and errors caused by patients' lack of expertise, including improper equipment operation and poor performance during expiratory airflow operation, which ultimately leads to incorrect treatment directions \cite{bonthada}. The latter uses the original volume-time curve to generate a flow-volume curve \cite{cosentino}, but the curve generated using this method often introduces noise, which may lead to model prediction errors.

Deep learning models that learn images or signals in low dimensions and are increasingly being used in disease detection tasks. S. Mei et al. proposed the SpiroEncoder method to represent high-dimensional raw spirogram signals in low-dimensional space and learn key physiological information \cite{mei}. S. Gadgil et al. proposed a Transformer-encoder-based method to capture the temporal dependency of signals from low-dimensional spirogram embedding sequences with learnable classification labels CLS and position encoding \cite{gadgil}. The Variational autoencoder (VAE) model achieves effective approximate reasoning for continuous latent variables by introducing a variational lower bound estimator, namely stochastic gradient VB (SGVB) \cite{kingma}. T. Yun et al. showed that VAE-encoders can capture important signals from spirograms that are missing from expert features, which are often helpful for disease diagnosis \cite{yun}. The performance of deep learning models such as encoders is often affected by the quality of the original data. At present, for the robust representation of images or signals, the BYOL method based on self-supervised learning \cite{grill} achieves robust representation of signals and achieves state-of-the-art performance when only using the same signal. C. B. Kumar et al. deployed the self-supervised method BYOL \cite{grill} and further proposed using the self-supervised representation learning (SSRL) method to train the encoder to learn the low-dimensional embedding of the ECG signal after data augmentation from the representation space, achieving higher accuracy in the classification task \cite{kumar}.

\begin{figure*}
\centering
\includegraphics[width=\textwidth]{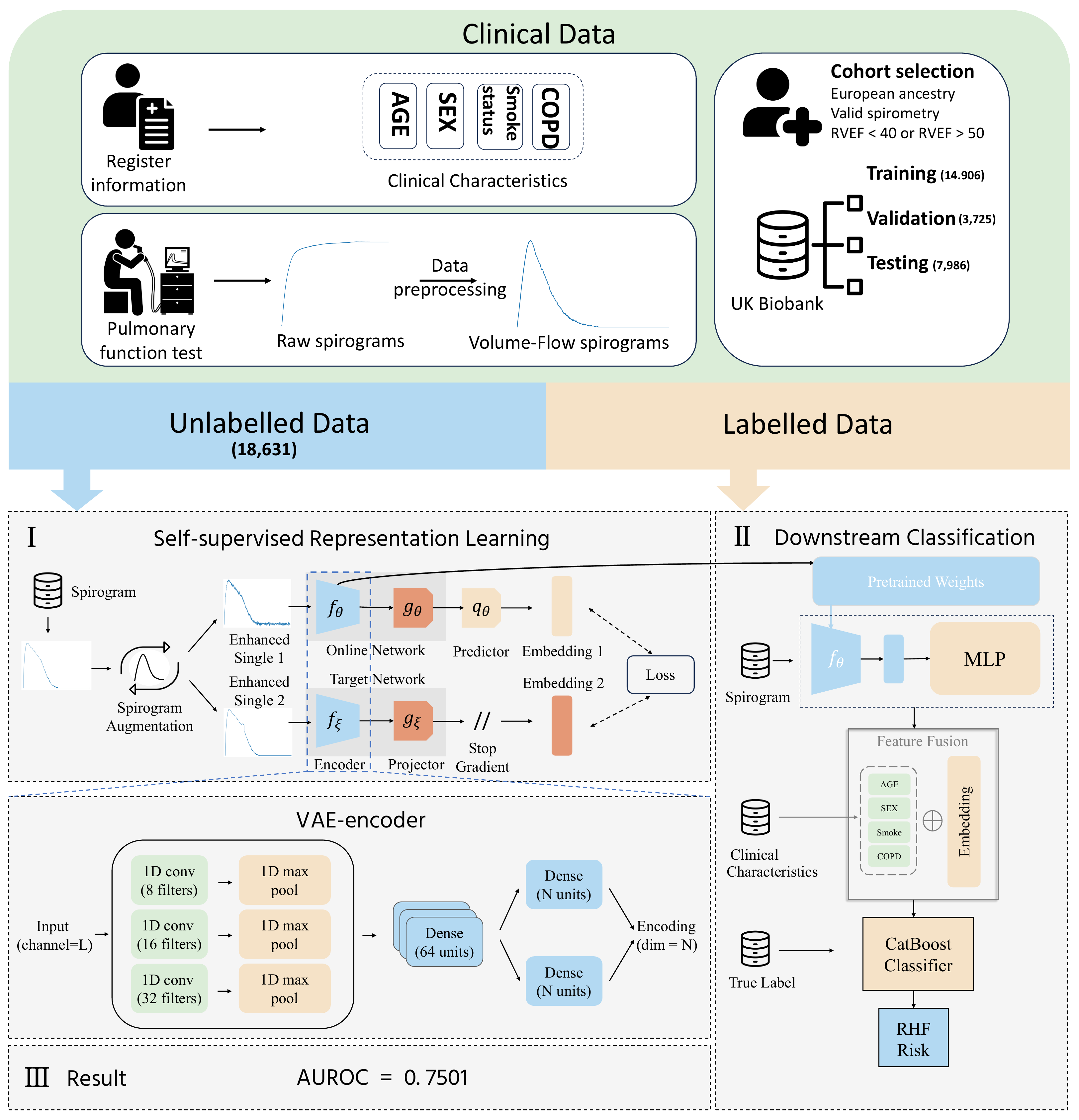}
\caption{\label{fig:overall_design}Overall Design. Our model utilizes clinical data including demographic information and raw spirograms collected from pulmonary tests. The Volume-Time spirograms are then generate to standardized Volume-Flow spirograms. The methodology is structured as a two-stage learning process using the UK Biobank dataset partitioned into training, validation, and testing sets. The proposed model proceeds in two stages: first, the SLSE network training stage, followed by the downstream classification task. The weight set of the online network is $\theta$, and the weight set of the target network is another different set $\xi$. Feature extractor using VAE-encoder. The SLSE network training phase completes the robust representation of the spirogram encoding, and the downstream classification task phase uses the CatBoost classifier with heterogeneous feature fusion as input to complete the prediction of patients with right heart failure.}
\end{figure*}

Therefore, we propose a self-supervised representation learning-based spirogram embedding (SLSE) method, which uses the information encoded from the spirogram to diagnose patients with RHF. The purpose is to screen out potential RHF patients from those who undergo spirometry tests and effectively increase the early intervention rate of RHF. The overall design of our method is shown in Figure \ref{fig:overall_design}. Our work has the following main contributions.

We propose to use spirogram as a new marker for diagnosing RHF. In clinical practice, patients suspected of having underlying lung disease will undergo spirometry tests. This helps to identify cor pulmonale patients who develop RHF at an early stage. We propose the SLSE method, which enables the encoder to learn a robust low-dimensional representation of the spirogram. The proposed model is divided into two training stages:

\begin{itemize}
\item In the SLSE network training stage, the VAE-encoder is trained to learn a robust low-dimensional feature representation from the data-augmented spirogram through a self-supervised representation learning method, and this process does not require any labeled data.
\item In the downstream classification task, the low-dimensional representation extracted from the original spirogram by the pre-trained VAE-encoder is fused with demographic information as the input of the CatBoost classifier, and finally predicts the early RHF patients.
\end{itemize}

We are the first study to use spirograms for the diagnosis of RHF patients. The model we proposed is simple to operate, does not require any expert knowledge, and has great potential for clinical application.

\section*{Method}

\subsection*{Dataset}
We use data from the UK Biobank in this study, which contains approximately 500,000 subjects. We use spirometry information from UKB field 3066, which contains 39,249 subjects and records the exhaled volume in milliliters recorded every 10 milliseconds. For volume-time, we first converted milliliters to liters on a volume scale, then, based on respiratory physiology and previous studies \cite{miller, cosentino}, we converted volume-time curves to flow-time curves using a finite difference approximation of the first derivative of time. Finally, flow-volume curves were generated using volume-time curves and flow-time curves, padded with zeros to a fixed length of 1000. We first filtered European individuals and only kept the first valid puff result. The dataset was then used for the next step of alignment with data from patients with RHF.

We use the method of determining whether or not there is RHF based on the right ventricular ejection fraction (RVEF) threshold to label the patients in the data with a binary RHF label. We use the RVEF measurement information from UKB field 24109, specifically from Instance 2, which contains 80,890 subjects. This field contains the RVEF measured using MRI. After referring to previous studies \cite{karlinsky, celant, mcDonagh, paul}, we decided to select \verb|RVEF=45%| as the threshold for RHF. Subjects with RVEF less than \verb|45%| are considered to have RHF and are labeled 1. Subjects with RVEF greater than or equal to \verb|45%| are considered to not have RHF and are labeled 0. It is worth noting that the existing method of measuring RVEF using MRI has unavoidable errors \cite{wang1, wang2}, which further leads to labeling problems. However, the unsatisfactory labeling will result in poor classification performance of the model. Therefore, we only selected the left and right extremes of the RVEF value in the population under the threshold of \verb|RVEF=45%| as the subjects of this study. Specifically, we screened out subjects with \verb|RVEF<40%| or \verb|RVEF>50%|, and after aligning the data with the subjects who was recorded in the spirometry test, 26,617 patients were left as the subjects of this study. In order to facilitate the observation of the prevalence of RHF in people of different age groups, we divided the people in the dataset into three age groups according to age: Youth (18-40), Middle (45-54), Elderly (55 +). It can be seen that the prevalence of right heart failure gradually increases with age.

Table \ref{tab:Demographic} shows the clinical characteristics of the UKBB dataset. These variables were collected at the time of the initial measurement, defined as the first available assessment in the UKBB. We use only age, sex, smoking status, and chronic obstructive pulmonary disease (COPD) status as inputs for model training. Throughout the data modeling process, age was an integer variable, while sex, smoking status, and other disease conditions such as COPD were binary variables. Among them, left heart failure (LHF) was a binary variable that was divided by the left ventricular ejection fraction (LVEF) with a threshold of 50. Specially, LVEF less than \verb|50%| are considered to have LHF and are labeled 1. LVEF greater than or equal to \verb|50%| are considered to not have LHF and are labeled 0.

The dataset is divided into \verb|70%| training set and \verb|30%| test set. During the model training process, the training set is used to train SLSE network, and in the downstream classification task, the training set is further divided into a validation set of \verb|20%| for validating the model. The test set is used for the final model performance evaluation.

\begin{table}[ht]
\centering
\begin{threeparttable}
\caption{\label{tab:Demographic}Clinical Characteristics}
\renewcommand{\arraystretch}{1.3}
\begin{tabular}{lccc}
\toprule
 & \textbf{RHF (n=244)} & \textbf{non-RHF (n=26,373)} & \textbf{p value} \\
\midrule
\textbf{Age} & 58.4 $\pm$ 7.0 & 54.9 $\pm$ 7.3 & 0.0000 \\
\textbf{Sex} &  &  &  \\
{Male} & 186 (76.2\%) & 11,652 (44.2\%) & 0.0000 \\
{Female} & 58 (23.8\%) & 14,721 (55.8\%) & 0.0000 \\
\textbf{BMI} & 27.7 $\pm$ 4.3 & 26.6 $\pm$ 4.2 & 0.0000 \\
\textbf{Obesity (BMI $\ge$ 30)} & 62 (25.4\%) & 4632 (17.6\%) & 0.0018 \\
\textbf{Smoke} & 134 (54.9\%) & 14,183 (53.8\%) & 0.7711 \\
\textbf{Hypertension} & 64 (26.2\%) & 4792 (18.2\%) & 0.0016 \\
\textbf{Diabetes} & 14 (5.7\%) & 788 (3.0\%) & 0.0207 \\
\textbf{CKD} & 6 (2.5\%) & 242 (0.9\%) & 0.0308 \\
\textbf{CHD} & 17 (7.0\%) & 828 (3.1\%) & 0.0013 \\
\textbf{LHF} & 178 (73.0\%) & 752 (2.9\%) & 0.0000 \\
\textbf{COPD} & 50 (20.5\%) & 3328 (12.6\%) & 0.0003 \\
\textbf{VHD} & 5 (2.0\%) & 203 (0.8\%) & 0.0582 \\
\bottomrule
\end{tabular}
\vspace{0.5em}
\begin{tablenotes}
\small
\item RHF = Right Heart Failure, BMI = Body Mass Index, CKD = Chronic Kidney Disease, CHD = Coronary Heart Disease, LHF = Left Heart Failure, COPD = Chronic Obstructive Pulmonary Disease, VHD = Valvular Heart Disease.
\end{tablenotes}
\end{threeparttable}
\end{table}

\subsection*{Model Architecture}
Self-supervised learning is an unsupervised learning technique that aims to enable models to learn useful information from unlabeled data. Existing studies have used self-supervised learning techniques in the field of image or physiological signal representation \cite{doersch, pmlr, xiao}. The self-supervised learning method BYOL achieves robust representation learning of signals, which achieves state-of-the-art model performance without using positive samples (signals of the same class) and negative samples (signals of different classes) \cite{grill}. The acquisition of physiological signals is subject to many factors and inevitably contains noise, which often leads to errors in using physiological signals to diagnose diseases. Recently, there have been studies focusing on extracting robust features from biological signals \cite{mei, kumar}. The SSRL method deploys the self-supervised learning method BYOL \cite{grill} and proves that BYOL can be used for representation learning of biological signals, and achieves robust representation learning of ECG signals \cite{kumar}. In medical prediction tasks, the use of heterogeneous features that fuse structured features and unstructured features can enhance the representation learning of patients \cite{zhang, seinen}. In disease prediction, demographic information can provide additional medical observations, which can help improve the performance of prediction models and reduce errors. CatBoost classifier \cite{prokhorenkova} is a decision tree that can better classify based on demographic information. In our model, CatBoost classifier \cite{prokhorenkova} takes heterogeneous features that combine spirogram embedding information and demographic information as input, and finally predicts patients with RHF. Our proposed model is shown in Figure \ref{fig:overall_design}. The proposed model consists of two training stages: the SLSE network training stage and downstream classification task.

The SLSE method focuses on learning robust low-dimensional feature representations from spirograms. In the SLSE method, we adopt the same deep learning framework as BYOL \cite{grill}, which uses two neural networks for learning: a target network and an online network. The online network is defined by weights $\theta$ and consists of three parts: encoder $f_{\theta}$, projector $g_{\theta}$ and predictor $q_{\theta}$, as shown in Figure \ref{fig:overall_design}. The target network is defined by another different set of weights $\xi$ and consists of two parts: encoder $f_{\xi}$ and projector $g_{\xi}$, as shown in Figure \ref{fig:overall_design}. The output of the target network is used as the label predicted by the online network. The target network parameter set $\xi$ is the exponential moving average of the online network parameter set $\theta$. After each batch of training steps, the parameter set $\xi$ performs the following update operations:

\begin{equation*}
\xi \leftarrow \tau \xi+(1-\tau) \theta \tag{1}
\label{eq:1}
\end{equation*}

\noindent where $\tau \in [0,1]$ is the target decay rate.

Given a training dataset $\mathbf{X}$, samples $x \sim \mathbf{X}$ uniformly sampled from $\mathbf{X}$, and two signal augmentation distributions $\mathcal{T}$ and $\mathcal{T^{\prime}}$. In each batch of training steps, two different signal augmentation techniques $t_1 \sim \mathcal{T}$ and $t_2 \sim \mathcal{T^{\prime}}$ are applied to the same signal $x$, respectively, to obtain two different augmented signals $x_{t1} \stackrel{\text{def}}{=} t_1(x)$ and $x_{t2} \stackrel{\text{def}}{=} t_2(x)$. The first augmented signal $x_{t1}$ is passed as input to the online network. In the online network, $x_{t1}$ is first passed to the encoder $f_{\theta}$, and then $x_{t1}$ is encoded into the representation space to form a representation vector $y_{\theta} = f_{\theta}(x_{t1})$. Next, $y_{\theta}$ is passed to the projector $g_{\theta}$ and generates a projection $z_{\theta} = g(y_{\theta})$ into another space. Similarly, the augmented signal $x_{t2}$ is used as the input of the target network, where the encoded vectors $y_{\xi} = f_{\xi}(x_{t2})$ and $z_{\xi} = g_{\xi}(y_{\xi})$ are generated by the encoder $f_{\xi}$ and the projector $g_{\xi}$, respectively. In the last stage of the online network, the predictor $q_{\theta}$ uses $z_{\theta}$ as input to produce $q_{\theta}(z_{\theta})$, which is the prediction of the target network's representation $z_{\xi}$. Finally, we normalize $q_{\theta}(z_{\theta})$ and $z_{\xi}$ to obtain $\bar{q}_{\theta}(z_{\theta}) \stackrel{\text{def}}{=} q_{\theta}(z_{\theta})/\|q_{\theta}(z_{\theta})\|_2$ and $\bar{z}_{\xi} \stackrel{\text{def}}{=} z_{\xi}/|z_{\xi}|_2$, respectively. Note that the online network and the target network are not symmetric because there is no predictor $q_{\theta}$ in the target network. We use the mean squared error between $\bar{q}_{\theta}(z_{\theta})$ and $\bar{z_{\xi}}$ as the loss function defined in equation \eqref{eq:2}:

\begin{equation*}
\mathcal{L}_{\theta,\xi} \stackrel{\text{def}}{=} \|\bar{q}_\theta(z_\theta) - \bar{z}_\xi\|_2^2 = 2 - 2 \cdot \frac{\langle q_\theta(z_\theta), z_\xi \rangle}{\|q_\theta(z_\theta)\|_2 \cdot \|z_\xi\|_2} 
\tag{2}
\label{eq:2}
\end{equation*}

We use Adaptive Moment Estimation (Adam) optimizer to minimize $\mathcal{L}^{SLSE}_{\theta, \xi}$ only for $\theta$ but not for $\xi$, as shown in Figure \ref{fig:overall_design} where the gradient updates to the target network are stopped. The dynamics of the SLSE network optimization process can be expressed as follows:

\begin{equation*}
\theta \leftarrow \text{Optimizer}(\theta, \Delta_{\theta}\mathcal{L}^{SLSE}_{\theta, \xi},\eta) \tag{3}
\label{eq:3}
\end{equation*}

\noindent where optimizer is an optimizer and $\eta$ is the learning rate. After SLSE training, we only keep the encoder $f_{\theta}$ and use it in downstream classification tasks.

In the downstream classification task, we use CatBoost \cite{prokhorenkova} as the final classifier to predict patients with RHF. Given the patient's demographic information $f_{demo}$ and the corresponding binary target label $y$, the pre-trained encoder $f_{\theta}$ takes the original spirogram signal $x$ as input to generate the spirogram encoding $f_{embed}$, which is then passed through the MLP to obtain a preliminary evaluation: $f^{\prime}_{embed} = \text{MLP}(f_{embed})$, and finally fused with the demographic information to obtain the predicted value $\hat{y} = \text{CatBoost}(f^{\prime}_{embed} \ \oplus \ f_{demo})$. In order to make the trained model have more stable detection performance, we use an ensemble learning method \cite{mahajan}. Specifically, we select the top 3 models that achieve the best performance of area under the receiver operating characteristic curve (AUROC) during training, then load these models when testing in the test set, and take the average of the predicted probabilities of these models as the final prediction result.

\begin{figure*}[ht]
    \centering
    \includegraphics[width=\linewidth]{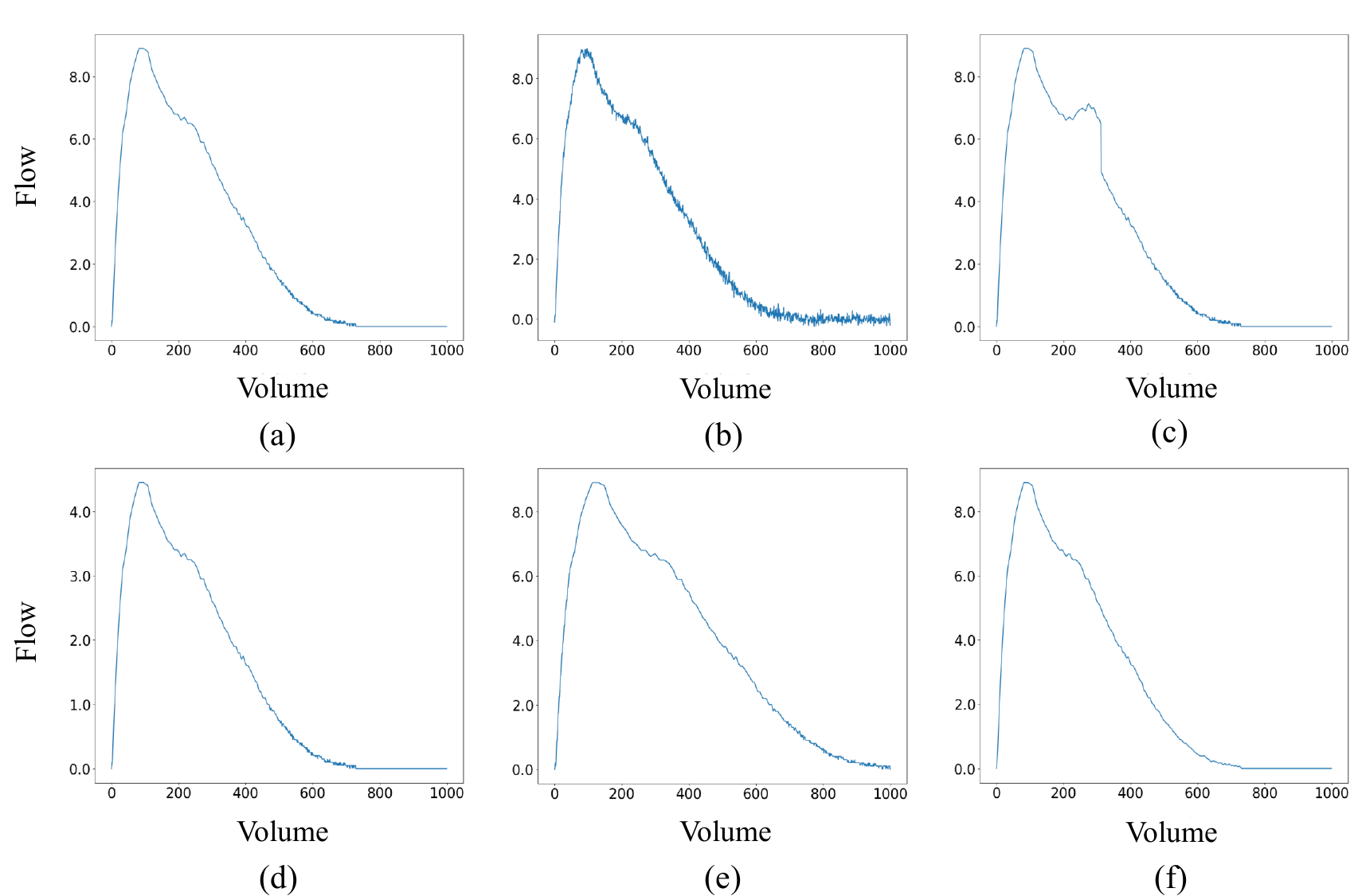}
    \caption{\label{fig:augmentation}Data augmentation method. (a) shows the original capacity-flow curve, (b) shows the curve after Gaussian noise addition, (c) shows the curve after a certain period of amplification after the peak, (d) shows the curve after vertical stretching, (e) shows the curve after horizontal stretching, and (f) shows the curve after downsampling.}
\end{figure*}

\subsection*{Encoder, Projector and Predictor Structure}
Using low-dimensional representations learned from biological signals by encoders has become an effective means of diagnosing diseases \cite{mei, gadgil, kumar}. Existing studies have suggested that VAE-encoders can learn signals from spirograms that are ignored by expert features, which are often helpful for the diagnosis of diseases \cite{yun}. Therefore, in our proposed model, we used the same VAE-encoder as in \cite{yun} as an encoder to provide information in spirograms that may be useful for diagnosing RHF. The architecture of the VAE-encoder is shown in the Figure \ref{fig:overall_design}. We choose the latent dimension 8 as the representation feature dimension of the VAE-encoder output. These low-dimensional feature representations can bring important signals ignored by expert features, so that the model can learn more effective information for the diagnosis of RHF from the spirogram.

Both Projector and Predictor use MLP networks here. In MLP networks, a linear neural network layer is followed by layer normalization, then RELU activation function, and finally another linear neural network layer. In the online network, the number of neurons in the linear neural network layer of Projector and Predictor is different.

\subsection*{Spirogram Augmentation}
In this part, we focus on simulating the equipment errors or patient operation errors that may occur in the clinical environment through data augmentation methods \cite{bonthada}, as well as the unstable curves \cite{cosentino} caused by the limitations of existing curve generation technology. The encoder can resist these errors, thereby learning a robust low-dimensional representation and enhancing the generalization ability of the model prediction. A total of five data augmentation methods are included (Figure \ref{fig:augmentation}). Gaussian noise: This method mainly simulates the unstable spirogram caused by the existing curve generation technology. Amplify a certain section after the peak of the curve: This method mainly simulates the irregular operation of some patients in the clinical environment who inhale after blowing. Horizontal stretching: This method mainly simulates the error of some patients in the clinical environment who inhale more gas through the nasal cavity during the blowing process, resulting in a larger exhaled lung volume. Vertical stretching: This method mainly simulates the error of too slow flow rate caused by a patient not blowing hard during the blowing process in the clinical environment. Downsampling: This method is mainly used to simulate the error caused by the machine with too low sampling frequency in the spirometry test. Here, the original capacity-flow curve is defined as $S(v),\{v=v_1,v_2,v_3, \dots, v_L\}$, where L=1000, indicating that the input capacity-flow curve has a fixed length of 1000. The specific implementation of the five data enhancement methods is as follows:

1) Gaussian noise: The original capacity-flow curve is $S(v)$, and the enhanced signal $S^{\prime}(v)=S(v)+N(v)$. Among them, $N(t) \sim \operatorname{Gaussian\left(\mu, \sigma^{2} \right)}$ is a random noise from a Gaussian distribution, and its length is the same as $S(v)$.

2) Amplification of a certain section after the peak: First, find the peak position $v_p$ from the original capacity-flow curve $S(v)$, and then define a mask function $M(v)$ using cosine smoothing to determine the position and length of signal amplification. The purpose of using cosine smoothing is to ensure that the amplification effect is natural and uniform. Here we define $M(v) =0.5 - 0.5 \cdot \cos\left(\frac{\pi \cdot (v - v_p - d)}{W-1}\right), \ v_p + d \leq t < v_p + d + W$, where $d$ is the random delay value after the peak, which is used to control the position of the signal amplification, and $W$ is the window size, which is used to control the length of the signal amplification. Finally, the amplified signal $S^{\prime}(v)=S(v) \cdot [1+M(t) \cdot (\gamma-1)]$ of a certain section after the peak, where $\gamma \in (1,\infty)$ is the amplification factor, which controls the degree of signal amplification.

3) Horizontal stretching: The original capacity-flow curve is $S(v)$, and the signal after horizontal stretching is $S^{\prime}(v)=S(\beta \cdot t)$, where $\beta$ is the stretching factor, which is determined by the ratio of the original signal effective blowing length $L_{vaild}$ to the signal fixed length L, $\beta=\frac{L_{vaild}}{L}$.

4) Vertical stretching: The original capacity-flow curve is $S(v)$, and the signal after vertical stretching is $S^{\prime}(v)=S(t) \cdot \alpha, \ \alpha\in(0,1)$, where $\alpha$ is the stretching factor.

5) Downsampling: First, apply a Butterworth \cite{wangboyou} low-pass filter $S_{f}(v)= \mathcal{B}(S(v), \frac{\alpha}{2})$ to the original capacity-flow curve $S(v)$, which ensures the quality of the signal when downsampling. Then, the intermediate signal $S_m(v^{\prime})=interp(S_f(v))$ is obtained by linear interpolation. In order to ensure that the signal has a fixed length L, the intermediate signal needs to be sampled to the target length L, and the final enhanced signal $S^{\prime}(v)=interp(S_m(v^{\prime}))$.

\begin{figure*}[!ht]
    \centering
    \includegraphics[width=0.8\linewidth]{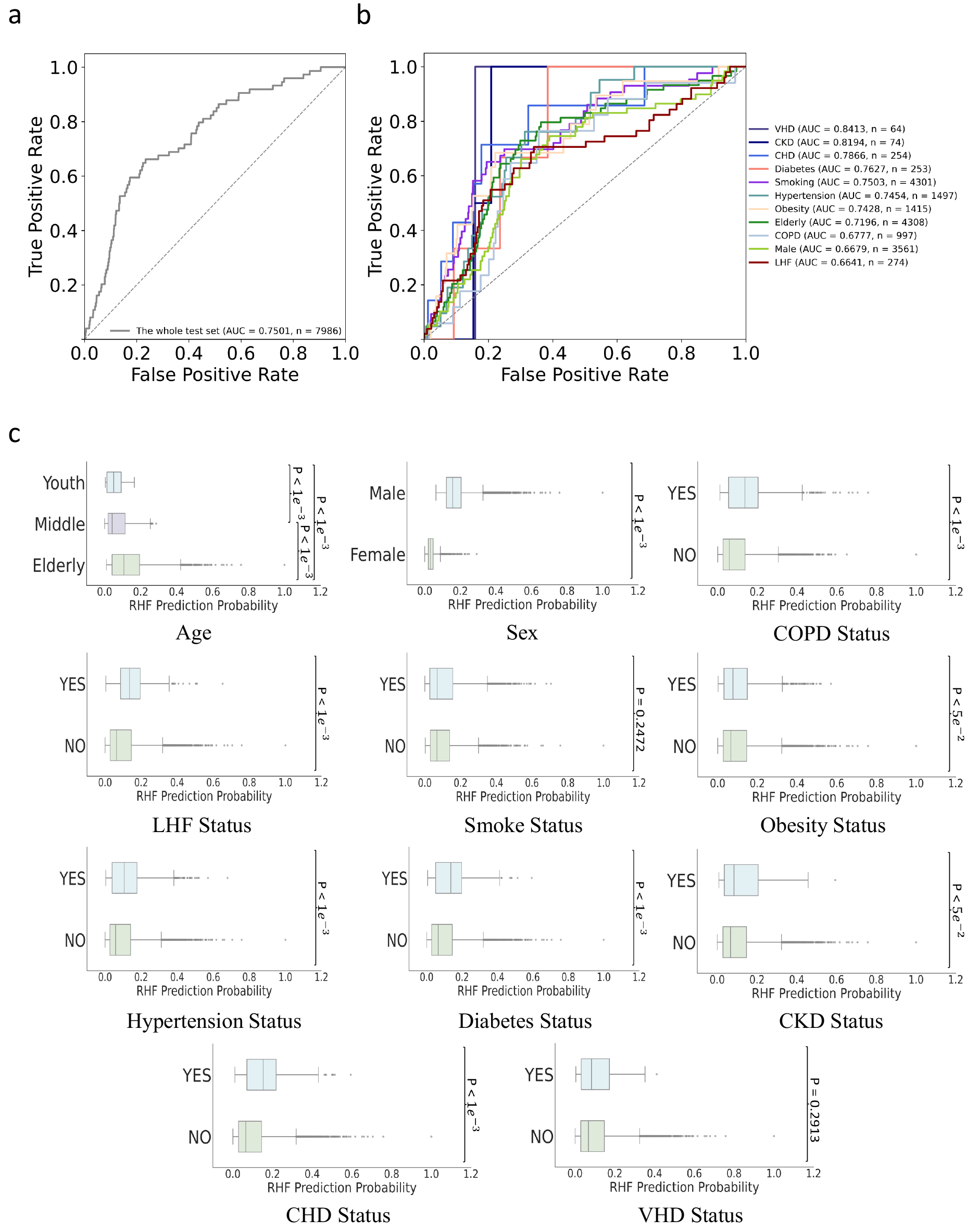}
    \caption{\label{fig:subgroup}Subgroup analysis. Subgroup analysis is performed on the test set for different subgroups to demonstrate the accuracy of the model classification in different subsets in the population. The subgroups involves age group, sex group, COPD status group, LHF status group, smoke status group, obesity status group, hypertension status group, diabetes status group, chronic kidney disease (CKD) status group, coronary heart disease (CHD) status group, and Valvular Heart Disease (VHD) group. \textbf{a} AUROC performance on the whole test set. \textbf{b} Subgroup analysis on AUROC performance. \textbf{c} Subgroup analysis on RHF prediction probability distribution. "YES" means that the patient has the disease or the condition, and "NO" means that the patient does not have the disease or the condition.}
\end{figure*}

\section*{Results and Discussion}
\subsection*{Performance Evaluation}
We choose area under the receiver operating characteristic curve (AUROC) as the indicator to evaluate model performance. For our task, AUROC is suitable for evaluating the performance of the model in a dataset with an imbalance of positive and negative samples. As shown in the Table \ref{tab:Demographic}, the number of healthy people in the dataset is much higher than that of patients with RHF, which is also in line with the actual clinical situation. Our model finally achieved AUROC=0.7501 on the test set, which shows that the model has good performance in predicting patients with RHF.

\subsection*{Subgroup Analysis}
We further explore the application of the model to other RHF risk factors. Specifically, we select subsets of RHF risk factors from the test set, each containing only one specific factor showing in Table \ref{tab:Demographic}. We then test the model's performance on these different risk factor subsets. Moreover, We conduct subgroup analysis on the model output results in terms of age, sex, smoking history and other risk factors for RHF to show the accuracy of our model in diagnosing different subsets in the population.

Ultimately, we find that the model achieves better AUROC performance on VHD, CKD, CHD, diabetes, and smoking subsets than when tested on the entire test set (Figure \ref{fig:subgroup}a), particularly achieving an AUROC=0.8194 on the CKD subset, and an AUROC=0.8413 on the VHD subset. This demonstrates the potential of our model for predicting RHF in CKD and VHD patients. For a excellent AUROC performance on this two subsets, it is possibly because CKD and VHD will increase circulatory load, which significantly exacerbates the symptoms of RHF when combined with it. As for the differences in subgroup distribution, Figure \ref{fig:subgroup}c CKD group shows that the RHF prediction probability distribution for patients with CKD is significantly broader than that for those without CKD, and the median prediction probability is also larger; the same applies to VHD subgroup (Figure \ref{fig:subgroup}c VHD group). The p-values indicate that the model has strong classification ability for the CKD subgroup but weaker classification ability for the VHD subgroup, suggesting that further validation with more VHD patient samples may be needed.

Age is a key factors in inducing RHF. From age group in Figure \ref{fig:subgroup}c, our model shows that the predicted probability of RHF increases with age, with the elderly (55+) group exhibiting higher probabilities and a broader range compared to youth (18-40) and middle groups (45-54). This is consistent with clinical expectations, as age is a significant risk factor for RHF. Similarly, the sex group in Figure \ref{fig:subgroup}c reveals that men have a higher predicted probability of RHF than women, consistent with studies showing a sex-based difference in RVEF, where men are more likely to have reduced RVEF \cite{zhao}. From the p-value, the model shows a strong ability to distinguish between the age and sex groups.

From other risk factors for RHF, patients with these conditions or diseases generally had a higher predicted risk of RHF (Figure \ref{fig:subgroup}c), which is consistent with clinical observations. From the p-value, the model also shows strong classification ability except for the smoke status subgroup.

\begin{figure*}[!ht]
    \centering
    \includegraphics[width=0.8\linewidth]{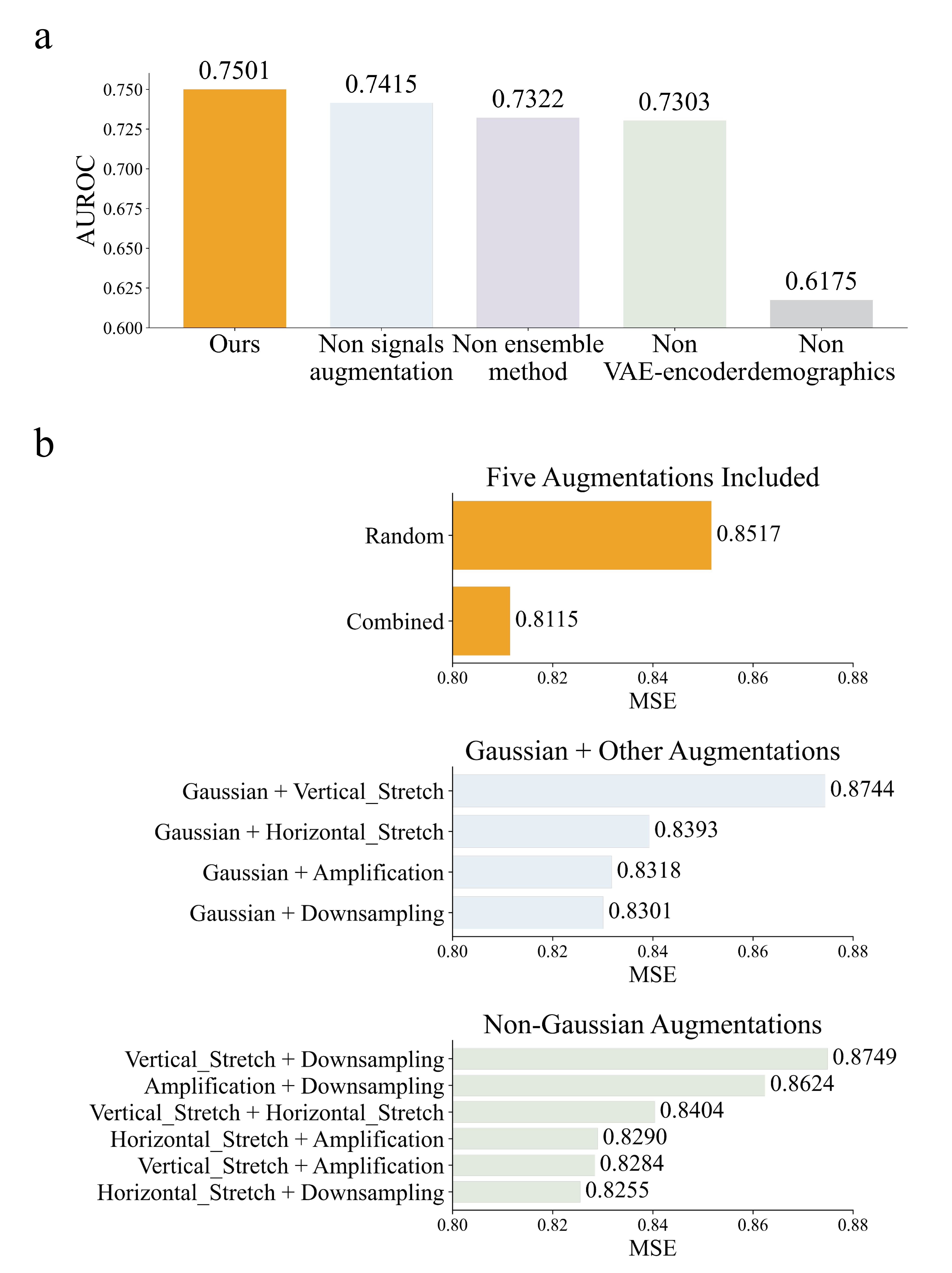}
    \caption{\label{fig:AUC&loss}Ablation experiment. \textbf{a} Ablation experiment of different modules. Comparing the AUROC performance between the methods after deleting each module and the original method on the same test dataset. \textbf{b} Ablation experiment of augmentation methods. It shows the loss obtained by using different combined signal augmentation methods. It can be seen that SLSE method enables the VAE encoder to learn a robust low-dimensional representation of the lung volume map with good anti-interference ability.}
\end{figure*}

\subsection*{Ablation Experiment}
In order to prove the effectiveness of each module in the model, we conduct ablation experiments on each module.

The specific operation is to delete only one specific module each time while keeping other modules unchanged from the original method, and conduct experiments to observe the performance of using only the remaining modules in the test set classification task. All methods are trained using the same dataset and evaluated on the same test set. From Figure \ref{fig:AUC&loss}a, it can be seen that without using the ensemble method, AUROC=0.7322, which is lower than the performance of the original method, indicating that compared with using only one best model for detection, using the ensemble method can make the model's detection results more stable. The method without using VAE-encoder only achieved AUROC=0.7303, which is significantly lower than the original method. This proves the effectiveness of the VAE-encoder module in the classification task of patients with RHF. VAE-encoder can effectively capture information related to RHF in the volume-flow curve of the spirogram. In the last experiment, during the SLSE network training stage, the input signal of the VAE-encoder was not augmented. This method finally achieved AUROC=0.7415, which is still lower than the original method. This shows that using signals augmentation can help the VAE-encoder resist noise interference from the original signal and obtain a robust low-dimensional feature representation. We also delete all demographic information used for training, using only spirogram embedding information and performing the classification task on the binary classification head. This method achieves an AUROC of 0.6175 on the same test set, which is significantly lower than the original method. This is consistent with previous research findings that in medical prediction tasks, using heterogeneous features that fuse structured and unstructured features can enhance the representation learning of patients.

In order to show that VAE-encoder can achieve robust feature representation in representation learning of spirogram, we conduct ablation experiments on five data augmentation methods.

One ablation method combines the five augmentation methods in pairs, for a total of ten combinations, and uses them as the data augmentation method for each SLSE network, and calculates the MSE loss of each method. Another ablation method combines the five augmentation methods together, also as the data augmentation method for the SLSE network, and calculates the MSE loss of this method.

From Figure \ref{fig:AUC&loss}b, it can be seen that whether one of the five methods is randomly selected as a data augmentation method or other augmentation methods are used, the MSE loss obtained by SLSE is almost the same and acceptable. This shows that the SLSE method achieves robust representation learning of spirogram and can resist errors and mistakes caused by improper patient operation, machines with different sampling frequencies, and existing curve generation technology in clinical practice.

\begin{figure*}[ht]
    \centering
    \includegraphics[width=\linewidth]{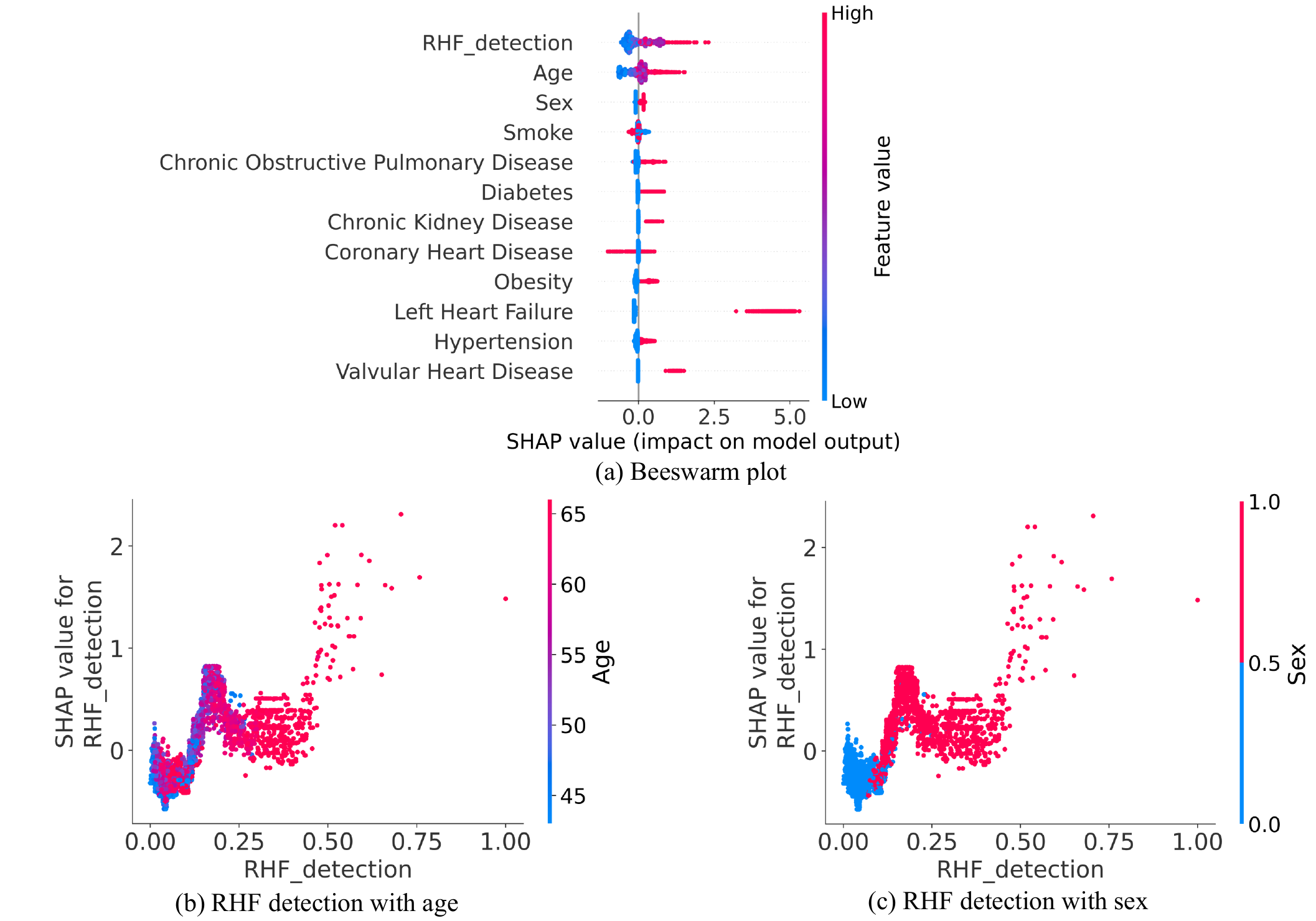}
    \caption{\label{fig:SHAP}SHAP analysis. (a) Beeswarm diagram. The darker the color, the greater the positive impact of the feature on the model prediction results. (b) Analysis of the dependence of RHF prediction results on age characteristics. The figure shows how age affects the model's prediction of RHF, showing that the elderly are more likely to be predicted as RHF. (c) Analysis of the dependence of RHF prediction results on sex characteristics. The figure shows how sex affects the model's prediction of RHF, showing that males are more likely to be predicted as RHF than females.}
\end{figure*}

\begin{figure*}[ht]
    \centering
    \includegraphics[width=\linewidth]{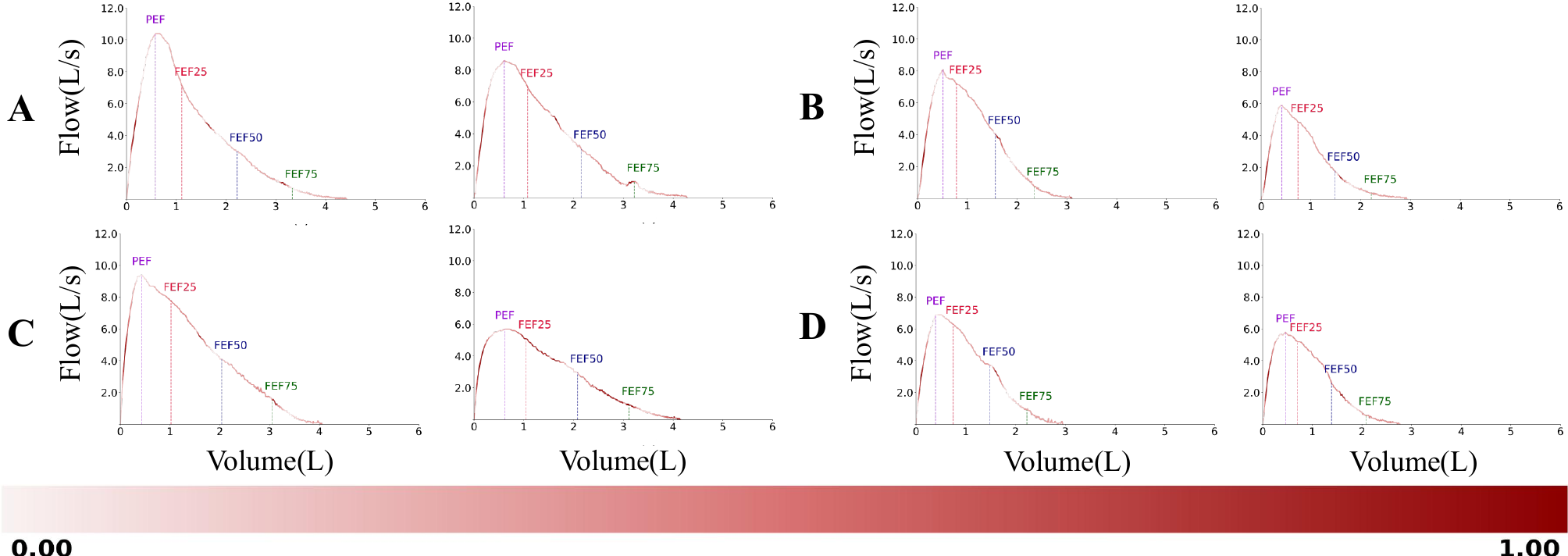}
    \caption{\label{fig:case_study}Case Study of patients with Right Heart Failure. Group A is True Positive, Group B is True Negative, Group C is False Positive, and Group D is False Negative.}
\end{figure*}

\subsection*{Interpretation Analysis}
To further analyze whether the model's diagnostic results are consistent with clinical significance, we perform SHAP analysis \cite{abdulalim}.

The beeswarm plot in Figure \ref{fig:SHAP}(a) reveals the relative importance of features to the model output. The plot shows that, overall, higher RHF risk scores have a greater impact on the model's predictions. Older age, male sex, obesity, and patients with COPD, diabetes, CKD, LHF, hypertension, and VHD have a strong positive impact on the model's output. The model's predictions are consistent with relevant research and medical observations, indicating that older age, male sex, obesity, and patients with the aforementioned diseases are predicted to be at higher risk of RHF.

We investigate the relationship between age and RHF risk value and the relationship between sex and RHF risk value though feature relationship dependency graphs. The results of Figure \ref{fig:SHAP}(b) shows the relationship between age and the risk of RHF. It can be clearly seen from the figure that the risk value predicted by the model is related to age. Being older have a strong influence on the model's prediction of positive samples. When the risk value exceeds 0.3, almost all patients are elderly. This is consistent with clinical significance. For RHF patients, the older they are, the greater the probability of having the disease.

Figure \ref{fig:SHAP}(c) shows the relationship between sex and RHF. There is a strong connection between the risk value and sex. Males have a strong influence on the model's prediction of positive samples. In addition, as the risk value increases, the proportion of male patients increases. When the risk value exceeds 0.3, almost all patients are male. This is consistent with clinical significance. For RHF patients, male patients are more likely to get sick.

An interesting finding shown in Figure \ref{fig:SHAP}(a) is that smoking appears to influence the model's prediction of negative samples. This may be because, when counting smoking history, subjects with only one or two smoking episodes are also considered to have a smoking history. Once or twice smoking is not necessarily a strong predictor of RHF. Therefore, when the model considers other factors (such as gender and age), smoking is considered a relatively unimportant factor, or even considered to have little to do with RHF.

Furthermore, Figure \ref{fig:SHAP}(a) appears to show that a significant portion of patients with CHD (red values) influence the model's predictions for negative samples (subjects without RHF). This means that these CHD patients are not detected as RHF by the model. One possible explanation is that not all CHD lead to RHF; only when severe myocardial ischemia occurs or after an acute myocardial infarction does myocardial function become significantly affected. However, the figure clearly shows that a portion of CHD patients still has some influence on the model's detection of positive samples. This is still consistent with clinical findings, that is, CHD is a risk factor for RHF.

Due to its "black box" nature, deep learning models are often not easily trusted by clinicians in the process and results of medical diagnosis. In order to explore the performance characteristics of RHF on the spirograph, this section adds an attention mechanism based on the VAE-encoder to study which parts of the volume-flow curve the model mainly represents, and conducts a case study analysis on the detection of right ventricular failure patients on the spirograph. Group A is RHF patients and diagnosed as RHF patients, Group B is non-RHF patients and diagnosed as non-RHF patients, Group C is non-RHF patients and diagnosed as RHF patients, and Group D is RHF patients and diagnosed as non-RHF patients.

As shown in the Figure \ref{fig:case_study}, the VAE-encoder pays attention to different segments in the volume-flow curve. Brighter colors indicate higher attention to the corresponding area, indicating that the area is a key segment, while darker colors indicate lower attention, indicating that the area is a non-key segment. The encoder automatically encodes these key segments.

From the Figure \ref{fig:case_study}, it can be seen that the encoder pays attention to the front, middle and end of the blowing stage, indicating that the encoder can roughly capture the key features of the patient in each blowing stage, so as to roughly grasp the patient's blowing situation. From the model's attention to the flow-volume curve, it can be seen that the model focuses on the segments before the peak expiratory flow rate ($PEF$), and the segments between $\text{FEF}_{25}$ - $\text{FEF}_{50}$ and $\text{FEF}_{50}$ - $\text{FEF}_{75}$. For some curves, the model pays special attention to the segment after $\text{FEF}_{75}$.

Regarding the curves of all disease types, the model pays special attention to the segment before the peak expiratory flow rate $PEF$. The possible explanation is that the forced exhalation process before reaching $PEF$ in healthy people is more regular and uniform, reflecting better lung function. In contrast, due to increased airway resistance and decreased lung compliance, this process may be more rapid or irregular in diseased people. This different performance of the segment before $PEF$ of the curve attracted the attention of the model.

For sick and healthy people, the encoder's attention to $\text{FEF}_{25}$ - $\text{FEF}_{50}$ is also different. Observing Figure \ref{fig:case_study}A, it can be seen that for sick people, the model's important attention is likely to appear between $\text{FEF}_{25}$ - $\text{FEF}_{50}$. For healthy people (Figure \ref{fig:case_study}B), the model's important attention is likely to be around or after $\text{FEF}_{50}$. This may explain the manifestation of RHF in the spirogram, that is, RHF has affected lung function to a certain extent, resulting in significant changes in mid-term expiratory function. This change is mainly manifested in impaired small airway function, which makes the more obvious and irregular flow rate drop characteristics appear earlier in the mid-term of exhalation, thus attracting special attention from the model. This finding provides a new observation perspective for clinical use of spirograms to assess the severity of RHF patients.

Interestingly, whether the second segment of the flow-volume curve falls before $\text{FEF}_{50}$ or after $\text{FEF}_{50}$ seems to have an important impact on the model's prediction results. From Figure \ref{fig:case_study}, the second segment of the focus for patients with RHF falls before $\text{FEF}_{50}$, and the second segment of the focus for patients without RHF falls after $\text{FEF}_{50}$. From Figure \ref{fig:case_study}C, due to the unstable performance of the patient during the inflation, the flow-volume curve was prematurely focused by the model in the $\text{FEF}_{25}$ - $\text{FEF}_{50}$ stage, and was mistakenly judged as a patient with RHF by the model. From Figure \ref{fig:case_study}D, the patient's right ventricular dysfunction may not have affected his lung function yet, which failed to attract the model's sufficient attention on the flow-volume curve. Therefore, the model's second focus segment fell after $\text{FEF}_{50}$, resulting in the patient not being detected as a patient with RHF.

\section*{Conclusion}
Looking back at this work, we first propose the use of spirogram as a marker for diagnosing right heart failure. In clinical practice, spirograms can be obtained through simple and widely available spirometry tests. The use of spirograms to diagnose right heart failure can help early detection and treatment of the disease. At the same time, we propose a self-supervised representation learning-based spirogram embedding method for right heart failure detection, which uses feature-robust spirogram encoding to diagnose right heart failure. Finally, the features of the spirogram encoding fused with demographic information are used to perform downstream classification tasks. The model achieves a performance of AUROC=0.7501 on the test set, indicating that the model has good disease diagnosis capabilities. This study lays the foundation for the use of spirograms for the diagnosis of right heart failure.

\section*{Data Availability}

This study utilized data from the UK Biobank, which can be accessed by approved researchers through the application process at https://www.ukbiobank.ac.uk. The UK Biobank has received ethical approval from the North West-Haydock Research Ethics Committee (REC reference: 21/NW/0157).

\section*{Acknowledgments}

Shenda Hong is supported by the National Natural Science Foundation of China (62102008, 62172018), CCF-Tencent Rhino-Bird Open Research Fund (CCF-Tencent RAGR20250108), CCF-Zhipu Large Model Innovation Fund (CCF-Zhipu202414), PKU-OPPO Fund (BO202301, BO202503), Research Project of Peking University in the State Key Laboratory of Vascular Homeostasis and Remodeling (2025-SKLVHR-YCTS-02). 

\section*{Author Contributions}

B.L., Y.Z., and S.H. conceptualized the study. Q.Z., Y.Z., and S.H. discussed the solution. B.L., Y.Z., and S.H. developed the methodology. B.L. developed the software. B.L., Q.Z., Y.Z., and S.H. validated the results. B.L., Q.Z., and Z.S. performed the formal analysis. K.L., S.G., and D.Z. carried out the investigation. Q.Z., S.G., and D.Z. were responsible for data curation. B.L. wrote the original draft. Q.Z., and S.H. reviewed and edited the manuscript. B.L., Z.S., and K.L. were responsible for visualization. All authors read and approved the final manuscript.

\section*{Declaration of Interests}
The authors declare no competing interests.

\newpage

\bibliography{reference}

\bigskip



\end{document}